\documentclass[runningheads]{llncs}
\usepackage{color}
\usepackage{graphicx}
\usepackage[caption=false]{subfig}
\usepackage{multirow}

\usepackage{amsmath}
\usepackage{algorithm}
\usepackage{algorithmic}

\begin{document}
	%
	
	\title{Knowledge Elicitation using Deep Metric Learning and Psychometric Testing}

	\titlerunning{Knowledge Elicitation using DML and Psychometric Testing}
	
	\author{Lu Yin \and
		Vlado Menkovski\and
		Mykola Pechenizkiy}
	\authorrunning{L. Yin et al.}
	\institute{{Eindhoven University of Technology, Eindhoven 5600 MB, Netherlands}\\
		\email{\{l.yin,	V.Menkovski,		m.pechenizkiy\}@tue.nl}}
	
	
	
	%

	%
	\maketitle              
	\begin{abstract}
		Knowledge present in a domain is well expressed as relationships between corresponding concepts. For example, in zoology, animal species form complex hierarchies; in genomics, the different (parts of) molecules are organized in groups and subgroups based on their functions; plants, molecules, and astronomical objects all form complex taxonomies. Nevertheless, when applying supervised machine learning (ML) in such domains, we commonly reduce the complex and rich knowledge to a fixed set of labels, and induce a model shows good generalization performance with respect to these labels. 
		The main reason for such a reductionist approach is the difficulty in eliciting the domain knowledge from the experts. Developing a label structure with sufficient fidelity and providing comprehensive multi-label annotation can be exceedingly labor-intensive in many real-world applications. In this paper, we provide a method for efficient hierarchical knowledge elicitation (HKE) from experts working with high-dimensional data such as images or videos. Our method is based on psychometric testing and active deep metric learning. The developed models embed the high-dimensional data in a metric space where distances are semantically meaningful, and the data can be organized in a hierarchical structure. 
		We provide empirical evidence with a series of experiments on a synthetically generated dataset of simple shapes, and Cifar 10 and Fashion-MNIST benchmarks that our method is indeed successful in uncovering hierarchical structures.

		
	\end{abstract}
	\section{Introduction}
	Supervised learning models specified as a map from the data-space to a fixed set of labels is the panacea of machine learning (ML) applications. However, our goal is often to `solve' a problem rather than to predict the labels. For example, consider a collection of texts or images -- we may ask people to tag them, but our goal is not necessarily to predict tags, rather to understand the `latent' taxonomy behind.  Or one can imagine a scenario of a medical diagnosis involving images (e.g.\ diabetic retinopathy) and a set of diagnostic images and a set of labels denoting the severity of the disease (e.g.\ `no disease,' `severity 1', ... `severity 4'). From the very beginning, the domain expert is forced to project her comprehensive knowledge on the five given discrete values. This limits our model both from its performance as it cannot learn from the full depth of knowledge as well from its interpretability as projecting to a fixed set of points further contributes to the 'black-box-ness' of our model.  In contrast, adding a more rich hierarchical structure to the data adds a significantly larger understanding of the finer subtleties of the data stemming from the relationships between the concepts in the domain, which provide us a taxonomy of different concepts relevant to decision making.
	
	The reason why so many solutions are formulated as mapping of datapoints (commonly high dimensional) to a fixed set of labels is that eliciting the knowledge from the domain and forming a training dataset is a difficult task. A data point can have a rich set of properties associated with it and eliciting an expert to provide an exhaustive annotation for each datapoint is prohibitively unscalable. 
	
	
	
	
	In this paper, we propose an approach that addresses this challenge by first developing a hierarchical representation of the data that facilitates more efficient annotation of data. Our method for knowledge elicitation consists of psychometric testing that captures the perceived relative distance of the datapoints for a given query (context) and a deep metric learning embedding algorithm. The embedded data are then structured in a hierarchical fashion that allows assigning properties and annotations to a large number of datapoints in a scalable way.  
	
	We evaluate the method empirically by showing we can capture the hierarchical knowledge structure in virtual responders for a given latent structure. Furthermore, using our method, we confirm the shape bias in the responses of a human participant.

	\begin{figure}[tbp]
		\centering 
		\includegraphics[width=0.8\columnwidth]{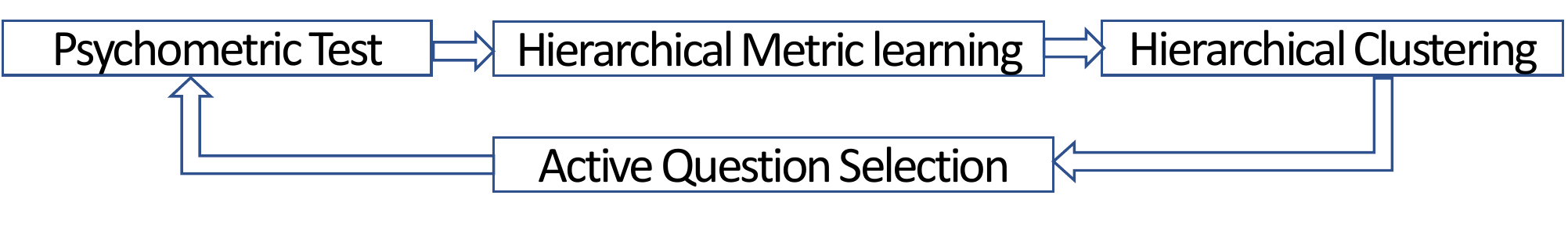} 
		\caption{Proposed hierarchy knowledge elicitation framework.} 
		\label{Whole_structure}
	\end{figure}

	\section{The proposed approach}
	We present an approach that consists of four components: psychometric testing, hierarchical metric learning, active question selection, and hierarchical clustering. The overall framework is shown in  Fig.~\ref{Whole_structure}.
	
	First, in order to capture the user's latent representation, we adopt a psychometric testing procedure~\cite{gescheider2013psychophysics} that relies on discriminative testing. The test collects information about the perceived relative differences to presented stimuli such as images or videos~\cite{son2006x,feng2014methodology}. Measuring the perceived difference with a psychometric test is significantly more accurate~\cite{gescheider2013psychophysics} then directly quantifying a perceived value. On the other hand, we also do not need to present the expert with target labels at this stage, so they can express their knowledge without mapping it to a predefined set of concepts. 
	
	The captured responses from the psychometric testing are used to develop a distance metric using a deep metric learning method. The goal of the metric learning component is to develop an embedding of the data in a space where distances reflect the representation of the expert. In other words, images that lie closer together are perceived as more similar than images that are further apart in the embedding. To achieve this, we extend existing metric learning techniques and introduce dual-triplet loss with an adaptive margin. Unlike triplet loss in~\cite{schroff2015facenet}, in dual-triplet loss, we do not distinguish from anchor image and positive image and apply a symmetrical loss structure to align with psychometric testing and fully take advantage of every sample in a loss function. We explained it in detail in the proposed approach section. 
	
	Psychometric testing offers many advantages. However, the number of all possible discriminative tests is $k$ combination of $n$, where $k$ is 3, and $n$ is the number of datapoints in the dataset. Asking all possible questions is typically not feasible. Nevertheless, to achieve a good embedding, we need only a small fraction of all possible questions. However, the quality of the embedding depends significantly on the selected questions as not all questions as equally informative.  To address this, we develop a question selection approach using a Bayesian-based active learning method that selects questions with high uncertainty and high utility.
	
	The embedding space is useful for many downstream tasks such as search and retrieval, but it also allows for efficient annotations. As now we have a metric to measure similarity, we can apply a label not only to a single datapoint but also to a region on the space covering multiple datapoints. Furthermore, if we can do this in a hierarchical fashion where labels at a different level of the hierarchy can be applied and then propagated to all lower levels. To achieve this, we combine the active question selection with a hierarchical clustering algorithm, such that we iteratively clusters and sub-clusters of the data to form a hierarchy and focus the question selection on sub-regions of the space. In this way, the approach starts first by forming the global distances and in turn organization of the data and then focuses on finer and finer differentiation as we go lower and lower in the hierarchy.
	
	\subsection{Psychometric Test}
	Psychometric testing is typically used to measure the subjectively perceived quantity of stimuli~\cite{gescheider2013psychophysics}. Different psychometric testing procedures are available, our method is a discriminative testing procedure, and a variation of the two-alternative forced-choice approaches. These methods typically present two alternatives to the participant that they are forced to choose from. One example of such set up would be to present two audio stimuli to a participant and ask which one is perceived as louder. To scale the perception of loudness, the experiment would consist of two pairs of stimuli, and the participant would need to answer which pair has a larger difference in loudness. Such experiments were also developed for estimating the quality of multimedia content~\cite{menkovski2012adaptive}
	. Another option is to present three stimuli and ask to discriminate between the relative difference between the two pairs formed by the three examples. However, many of these questions can be ambiguous, as the images may seem to have similar distances, or they may present different aspects that are not directly comparable in a pair-wise fashion~\cite{hellinga2019hierarchical}.
	
\begin{figure}[htbp]
		\centering 
		\includegraphics[width=0.8\textwidth]{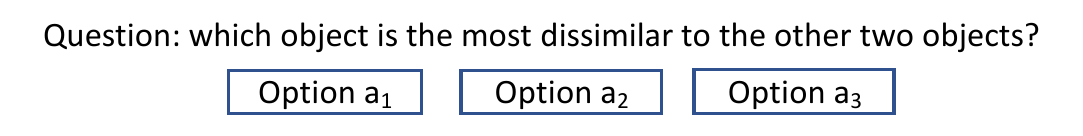} 
		\caption{Three-alternative-forced choice interface.} 
		\label{3AFC}
	\end{figure}

	To deal with this, we adapt the psychometric testing such that the participants are presented three objects $a_1, a_2, a_3$, and are forced to choose the most dissimilar one among them (Fig.~\ref{3AFC}). By carrying out this simple test, we can extract hierarchical knowledge from the data by simple ternary decision. The differences between these three objects are expressed as distances in our neural network model. The choice is based on the annotator's personal perception. Therefore, different annotators may have different choices when facing the same question, and different hierarchical trees are created. We examine in detail how the method deals with these conditions in the experiment section.
	
	\subsection{Hierarchical Metric Learning} 
	
	To successfully project the data in an embedded space that captures the latent representation of the expert, we train a model using a variation of the triplet loss function. The triplet loss given in Eq.~\ref{o_loss} and its training process was introduced in~\cite{schroff2015facenet}, consists of training using triplet of datapoints ($x_a^i,x_n^i,x_p^i$),
	
	
	\begin{equation}
	L=\sum_{i=1}^N \left[   d(x_{a}^i,x_{p}^i) - d(x_{a}^i,x_{n}^i) ,+m \right]_+ 
	\label{o_loss}
	\end{equation}
	where $N$ is number of the possible triplets,  $d(x_{a}^i,x_{p}^i)=||(f(x_a^i)-f(x_p^i)||_2^2$ and $d(x_{a}^i,x_{n}^i)=||(f(x_a^i)-f(x_n^i)||_2^2$, $f(z)$ is the representation of the data point $z$ in embedding space and $f$ is our model or more precisely $f_\theta$, where $\theta$ represent all the model parameters. hinge function $\left[ \cdot \right]_+$ indicate $max[0,\cdot]$ 
	When the data is annotation with a fixed set, metric learning is typically implemented such that the triplets take the following roles: $x_a^i$ is the anchor image, which has same label with positive image $x_p^i$ and different label with negative image $x_n^i$.
	Therefore, the triplet loss produces a loss value when two images with the same label are further apart than two images with different labels.
	
	Note that the form of triplet loss in Eq.~\ref{o_loss} naturally fits with how we defined the 3AFC psychometric test. Based on the participant's response, we can select the most different image as the negative sample , and the other two images as positive and anchor images. 
	
\begin{figure}[htbp]
		\centering
		
		\subfloat[ ]{\includegraphics[width=0.4\textwidth]{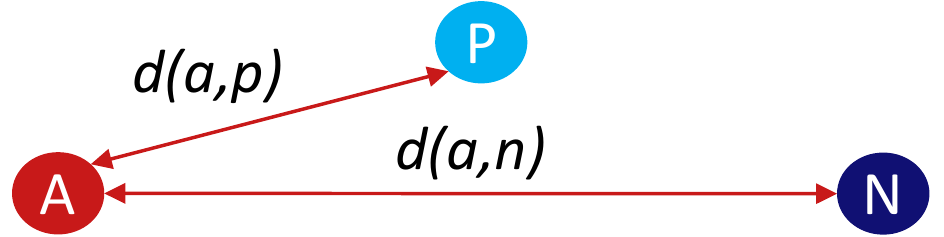}}
		\subfloat[]{\includegraphics[width=0.4\textwidth]{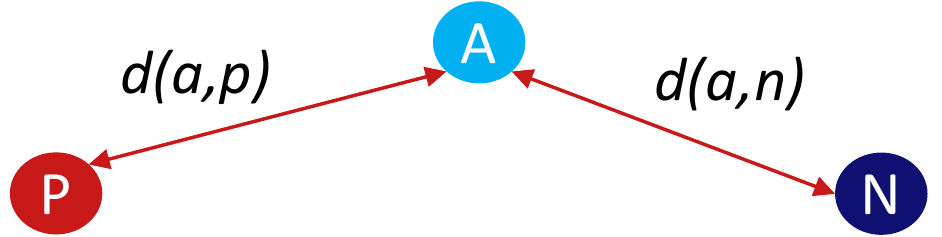}}
		\caption{(a) and (b) are different cases of choosing the anchor image. A represent the anchor image, N denotes negative and P for positive. Negative is chosen in 3AFC test and fixed, and we need to select the anchor image from the rest two. If $d(a,n)-d(a,p)$ is too large there will be limited gradients, and if it is larger than a margin value there will be no gradient.  Since in case (a) the anchor is farther away from the negative image than in case (b), it is more likely that case(a) results in limited or zero gradients while case (b) results in more gradients.}
		\label{2_case}
	\end{figure}

	However, the anchor and the positive image are not interchangeable in the triplet loss term as the distance of the anchor to the positive and negative image are compared. One can imagine a scenario where selecting one of the images as an anchor results in a gradient for the model and selecting the other one does not, or results limited gradient, as depicted in Fig.~\ref{2_case}.
	
	To deal with this, we propose a dual-triplet loss function in which there is no specific differentiation between the anchor and the positive image but rather define two positive and one negative image. The participant gives an answer about which image is the furthest from the other two, and set it to be negative. The dual-triplet loss is defined as,
	
	\begin{equation}
	L=\sum_{i=1}^N \left[d(x_{p1}^i,x_{p2}^i) - d(x_{n}^i,x_{p1}^i) ,+m_a^i \right]_+ +  \left[ d(x_{p1}^i,x_{p2}^i) - d(x_{n}^i,x_{p2}^i) ,+m_a^i \right]_+  \label{n_loss}
	\end{equation}
	where, $x_{p1}^i$ and  $x_{p1}^i$ are two positive images, and $x_{n}^i$ are the negative image chosen by annotator during 3AFC test, $N$ is the number of sampled triplets, $m_a^i$ is an adaptive margin which will be explained later. Now negative image is compared with each of the rest two images, and both situations in Fig. \ref{2_case} are considered. Every sampled is used twice in one triplet function to produce more gradients and to prevent zero gradients situation happen.
	
\begin{figure}[tbp]

		\centering 
		\includegraphics[width=0.6\textwidth]{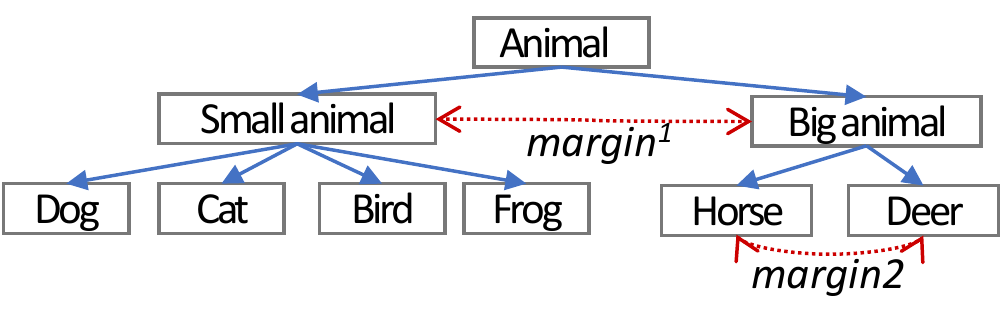} 
		\caption{Questions about small animal and big animal are more general and have a larger diversity than questions about horse and deer. So $margin^1$ should be larger than $margin^2$.} 
		\label{Margin}
	\end{figure}

	The triple loss margin has a significant impact on the performance of the metric learning model. Different methods have been developed to determine the value of the margin.  Some of these approaches~\cite{ge2018deep} propose using a margin value that adaptively changes during the training process. Since we collect the responses in a hierarchical manner, the margin  needs to adapt accordingly, from large to small, as the questions become more focused.  
	For example, if the responses are collected over the whole dataset, the margin needs to be appropriately large, and if we descent and become more focused in the testing, the margin needs to adapt and be smaller, as shown in Fig.~\ref{Margin}.
	Specifically we compute the  margin $m_a$  value based on the diversity between the three samples in a triplet $i$ and is defined as,
	
	\begin{equation}
	m_a^i=m_h^i+\gamma d_H^i\label{adaptive_margin}
	\end{equation}
	where $m_h^i$ is a constant value that guarantees $m_a^i$ is not zero and $\gamma d_H^i$ is the adaptive part. $\gamma$ is a hyper-parameter that we set during training. 
	$d_H^i$ is a diversity factor of the triplet $i$. We calculate the diversity factor at a node $i$, by computing the average distance of the centroids of all sub-clusters of the cluster associated with the node $i$, as given in Eq.~\ref{dh}, 
	
	\begin{equation}
	d_H^i=\frac{1}{{n^i}^2-{n^i}}\sum_{c_p,c_k\in L^i} ||c_p - c_k||^2_2
	\label{dh}
	\end{equation}
	where $L^i$ is a node where question $i$  is sampled from in the hierarchy. $n_i$ is the number of child nodes in $L^i$, $c_p$ is the center of cluster $p$, and $c_k$ is the center of cluster $k$.

	\subsection{Active Questions Selection } \label{AQS} 
	Given a data set containing $B$ images, there are  $\binom{B}{3}$ potential questions for the annotators to answer.  It is impractical to answer all of them, and randomly selecting the questions is suboptimal with respect to the efficiency of the training process. To address this, we develop an active question selection scheme.
	
	
	
	
	
	We start by randomly selecting $m$ questions for the annotators to answer. With the answers, we train the model $M$, create a hierarchical tree $H$, and construct a set of knowledge pool containing the answered questions $D$. The method iteratively repeats the process of question selection and update of the model $M$ as well as the hierarchical representation of the data $H$. 
	To select the questions for the following iteration, our method takes two steps. In the first step, a set of questions is proposed uniformly sampled from each node in the hierarchy $H$, and in the second step, we reject some of the proposed questions for which we do not expect high utility. 
	
	
	
	By sampling from each level of the hierarchy in the first step, we maximize the probability that the model will receive information both about the global distribution of the data as well as the specific differences at a finer level of detail. In contrast, randomly selecting questions is sufficiently less efficient in achieving the same goal. 
	
	
	Then, we use a pool-based active-learning question rejection scheme. We consider not only the uncertainty of the questions but also its utility, that is, select questions with high uncertainty and high variance. High uncertainty means annotators are not so sure which option to choose, thus the question could be informative, and its answer has a higher probability of introducing information to our model. High variance means similar questions have been answered many times, and this is to avoid ambiguous questions (i.e., similar questions have occurred too many times, but annotators are still not sure how to answer).
	
	In order to evaluate the uncertainty and utility, we need to calculate the distribution of answering possibility by Bayes' theorem. For a given question $q_i$ consists of 3 samples,   $\theta=(\theta_1,\theta_2,\theta_3 )$ denote the possibility of choosing $a_1,a_2,a_3$. We assume it follow dirichlet distribution $ Dir(\alpha_1,\alpha_2,\alpha_3) $, and it's density can be written as,
	
	\begin{equation}
	p(\theta|\alpha)=\frac{\Gamma(\alpha_1+\alpha_2+\alpha_3)}{\Gamma(\alpha_1)\Gamma(\alpha_2)\Gamma(\alpha_3)}\prod_{i=1}^3 \theta_i^{\alpha_{i-1}} (\theta_i\geq0; \sum_{i=1}^3\theta_i=1)
	\end{equation}
	where  $\Gamma$ denotes the gamma function, $\alpha=(\alpha_1,\alpha_2,\alpha_3)$. $\alpha_1=\alpha_2=\alpha_3=1$ at beginning, means the possibility of choosing $a_1, a_2$ or $a_3$  is equal.  The distribution is updated based on the previous answered questions as,
	\begin{equation}
	p(\theta|\alpha,D)=p(\theta|\alpha+m)
	\end{equation}
	where $D$ is the set of answered questions.  We define the questions sampled from neighbour questions of $a_1,a_2,a_3$ as $q_i$'s similar questions, and count the times of choosing $a_1,a_2,a_3$ when facing similar questions of $q_i$ in $D$. The counting number are defined as $m =(m_1,m_2,m_3)$.
	
	Now we evaluate the uncertainty by the maximum of the expected possibility of choosing any of  $a_1, a_2, a_3$. It does not matter which one we choose, because if any of them has a high possibility to be chosen, we can conclude that the model is confident about this choice based on the previous answers, and this question has less uncertainty. So if the maximum expected possibility $e(q_i)$ of choosing $a_1,a_2$ or $a_3$ is higher than  $s_e$ we will reject this question. Similar, we can evaluate the utility by the sum of possibility variance $var(q_i)$. If it is too high, we can interpret it as after facing similar questions many times, the annotator still has an ambiguous answer, so we also reject it.
	
	\subsection{Hierarchical Clustering}
	After answering enough actively selected questions, we embedded the data using the deep metric learning method such that its semantic hierarchical information represented by the Euclidean distance in that space. The hierarchical structure is extracted by a hierarchical clustering algorithm. Generally, there are two categories of methods for this task, a top-down approach called ``Agglomerative" and a top-down approach called ``Divisive"~\cite{rokach2005clustering}. 
	Since we have to decide the threshold very carefully if applying the ``Agglomerative" approach, in our experiment, we choose the" Divisive" approach K-means to perform the divisive cluster, and Silhouette Coefficient is applied to help us choose $K$. 
	
	\section{Related work}
	
	Our work is related to three lines of research (1)~Building hierarchical image structure (2)~Deep metric learning (3)~Informative samples mining.
	
	\textbf{Building hierarchical image structure}
	There are two branches explored in building a hierarchical image structure. One is with the help of language information. Such as WordNet is used to build a semantic hierarchic classifier~\cite{marszalek2007semantic}. However, the language-based similarity is not always what we want. When comparing whitefish, goldfish, and cat, the former two are closer in word semantic because they are all fish,  but this could not suit our needs if we want to build a hierarchical pet data set based on their cuteness. On the other hand, hierarchies could also be formed based on image features that are extracted by SIFT, HOG~\cite{wigness2018efficient}, or deep convolutional network~\cite{zheng2017hierarchical}. While able to group images based on their visual similarities, the hierarchies are not build based on the user's perception or knowledge, so it is also difficult to cater to a special user's preference (e.g.\ expert domain).

	\textbf{Deep metric learning} Deep metric learning is a data-driven approach to learn the measurement of similarity. It learns a nonlinear mapping for the original data to an embedding. Contrastive loss~\cite{hadsell2006dimensionality} is proposed to encourage the distance of positive paired samples be closer than negative samples of a margin of $m$. It is extended to triplet loss~\cite{schroff2015facenet}, quadruplet loss~\cite{chen2017beyond}, N-paired loss~\cite{sohn2016improved} where three, four, $N$ samples are used in loss function, and lifted structure loss~\cite{oh2016deep} which take advantage all the positive and negative pairs in a mini-batch.
	Although yield promising results in distances based computer vision applications such as face recognition~\cite{schroff2015facenet}, person re-identification~\cite{chen2017beyond}, and few-shot learning~\cite{oreshkin2018tadam}, their studies focus on the flat distance which has a weaker hierarchical distances representation ability.    
	
	\textbf{Informative samples mining} 
	For pair-based deep metric learning tasks, there are $O(N^n)$ potential tuples for training, $N$ is the sample number in training set, and $n$ is sample number in a tuple. It is  impractical to train with all of them due to the GPU memory or train time concern. Besides, a lot of them are redundant or less informative and contribute little gradients during training. A semi-hard mining scheme is used in FaceNet~\cite{schroff2015facenet} to select informative tuples online inside a mini-batch.  Harwood~\textit{et al.} considers the global structure in their smart mining~\cite{harwood2017smart} for producing effective samples with a low computational cost.  A distance weighted sampling strategy~\cite{wu2017sampling} is proposed to select samplings uniformly based on their relative distance. While these strategies boost the model performance by learning a more discriminative flat representation of each category object, our work aims to map data to an embedding space where a hierarchical structure is also considered. Therefore, a hierarchical sampling method with Bayesian-based active learning is proposed in our paper to better cope with our needs.

	\section{Experiments}
	
	As a first in our empirical evaluation, we aim to confirm that our approach can elicitate the latent hierarchical structure of a single participant. To remove any other sources of variation, we design an experiment with a virtual participant that always responds according to its given latent hierarchical model. In this manner, we can objectively evaluate the precision of the elicitated model against the one given to the virtual participant. 
	
	Next, we expand on this using multiple virtual participants that have partial agreements of their latent models. In large scale human experiments, we do not expect that the internal hierarchical representations of people will fully agree with each other. Therefore, in this analysis we aim to assess whether our approach can extract the common hierarchical model of a group of participants that have models with partial agreements. Towards this we simulate multiple individual participants and test whether the model can elicit common hierarchy accordingly.
	
	Then, we evaluate with a human participant. The goal of this experiment is to evaluate how well our method can deal with the additional variability involved in using responses from a person. In this setting, the additional challenge is that we cannot be precisely sure of the human's latent hierarchical model of the data. To deal with this, we specifically design an experiment that would confirm a well understood psychological bias that humans have, specifically the shape bias~\cite{landau1988importance}. The shape bias states that shapes are more important than other properties of an object, such as colors when we aim to discriminate between them. We develop an image dataset that allows us to test is we can confirm this bias in hierarchical models that we elicit from the responses. 
	
	Finally, we develop an experiment on a natural image dataset, specifically the Fashion-Mnist~\cite{xiao2017fashion}. Our aim here is to test in natural settings. In this case, our evaluation is limited to the flat set of labels that the dataset comes with.

	In all experiments, two evaluation metrics are considered: (1)\textit{The accuracy of each cluster}.  Every cluster in the hierarchy should share a similar concept. We tag a cluster by the majority shared concept in it, and calculate the accuracy of that node by counting the number of correctly assigned images and dividing by overall image number in that node. (2)\textit{The dendrogram purity between the extracted structure and the ground-truth.}  Since we want the hierarchy able to cater for special users, we can only claim a hierarchy is good when structure formed according to the ground-truth, which is the user's latent perception.  That can be evaluated by dendrogram purity~\cite{heller2005bayesian}. Rather than the flat cluster purity, it is a holistic way to evaluate the whole hierarchical structure tree. The value of it ranges from 0 to 1, 1 means every sample fits perfectly for the ground truth and on the contrary for 0. Note that dendrogram purity is only applicable when the ground-truth hierarchy is given in advance, so we only calculate it in simulation experiments, and as we are applying hierarchy clustering in this paper, every node of the hierarchy is a cluster.
	

	\subsection{Single Virtual Participant on Cifar 10 }\label{sec_singel_cifar_10}  
	
	
	With our proposed hierarchical knowledge elicitation method, we experiment on a snipped tiny Cifar 10. The data set containing 1000 images sampled randomly from the original Cifar 10 with 100 images in each category. 
	
\begin{figure}[htbp]

		\centering 
		\includegraphics[width=0.94\textwidth]{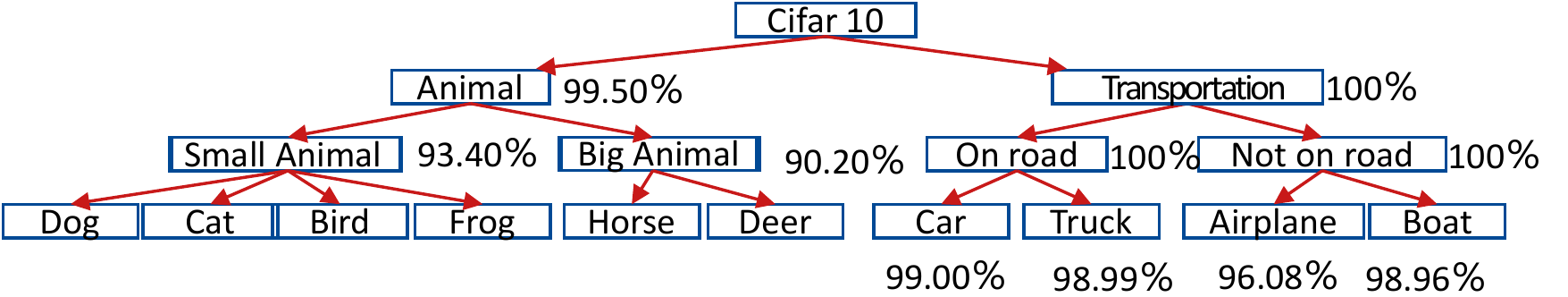} 
		\caption{Latent given hierarchy of a single virtual participant and the extracted result.} 
		\label{Simulated_single_participants}
	\end{figure}

	Since we might not be able to know a user's latent perception before the experiment, in order to evaluate how well our extracted structure matches the user's latent perception, we pre-define a specific hierarchy (see Fig.~\ref{Simulated_single_participants}) and run the experiment by simulation. A virtual participant gives perfect responses based on the given hierarchical structure. 1000 questions are simulated for model training with a fixed margin of 0.4 at the beginning. 600 more questions are sampled per iteration and used for training with adaptive margin in the following 5 iterations. 
	
	After collecting 3400 automatically respond triplets,  we show our results in Fig.~\ref{Simulated_single_participants}. We can see that our model can elicitate a clear hierarchical structure, which matches the given one well. Note that the small and big animals do not split down to individual animal nodes. That may because animal images are similar to each other, and it is not easy for the model to get a clear cluster with limited animal training triplets. We calculate the accuracy of the extracted nodes which are all above $90\%$.
	
	\subsection{Multiple Virtual Participants on Cifar 10 }\label{sec_mutiple_cifar_10}
	
\begin{figure}[htbp]
		\setlength{\belowcaptionskip}{-0.5cm} 
		\centering
		
		\subfloat[Virtual participant 2 and the extracted result.]{\includegraphics[width=0.94\textwidth]{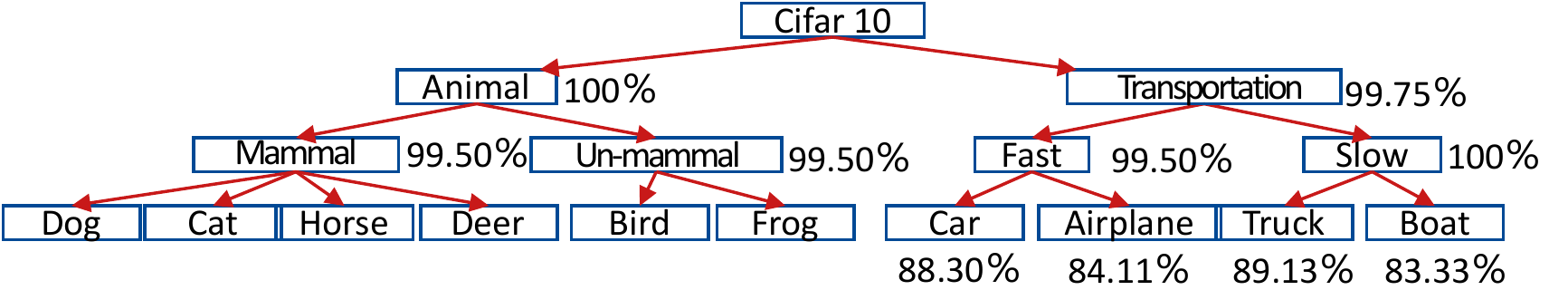}}\quad
		\subfloat[Virtual participant 3 and the extracted result.]{\includegraphics[width=0.94\textwidth]{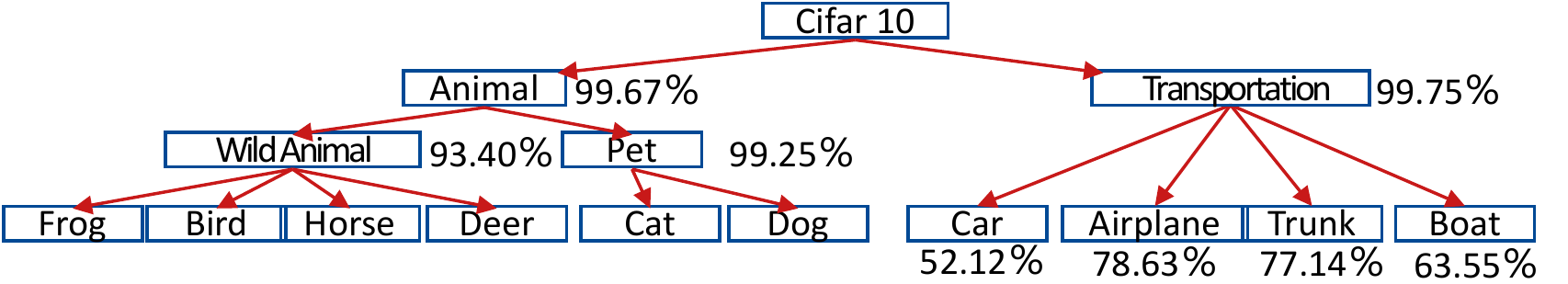}}\quad
		\caption{Latent given hierarchy of multiple virtual participants and the extracted result.}
		\label{Simulated_multiple_participants}
	\end{figure}

	We further evaluate our method in multiply-users situations when they have partial disagreements. Two more participants are simulated with different pre-designed hierarchies, as shown in Fig.~\ref{Simulated_multiple_participants}. We follow the same experiment setup with single-user situation in Section~\ref{sec_singel_cifar_10}.
	
	We illustrate our results in Fig.~\ref{Simulated_multiple_participants}, where we can see our model is able to extract varying hierarchical structures according to different given latent hierarchies.  Individual animal nodes still can not be clustered well due to their higher similarity and limited training samples.
	
	We also mixed the collected triplets from all three users, and extract a general hierarchy from all the triplets (See Fig.~\ref{Simulated_multiple_participants}). Due to enough training triplets, individual animal nodes got extracted, and other nodes get better accuracies compared to single participant situations. We can also notice that both shared perception nodes (10 original Cifar 10 classes, animal and transportation) and individual perception nodes (big animal from participant 1, un-mammal from participant 2, pet from participant 3) can be seen in the structure. 
	Therefore, our model can elicit a common  structure that represents both shared and individual knowledge of all participants.

	\begin{figure}[htbp]
		\centering 
		\includegraphics[width=0.94\textwidth]{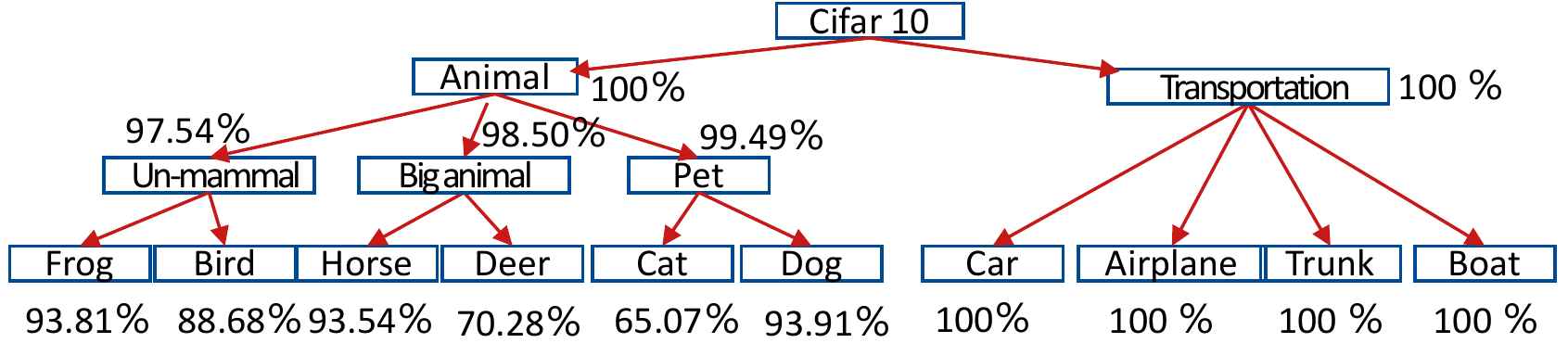} 
		\caption{Extracted common structure.} 
		\label{overall}
	\end{figure}
	
	\subsection{Real Participant on Synthetic Geometric Shape Data Set}\label{sec_real_shape}  
\begin{figure}[htbp]
		\centering
		\subfloat[The first three layers of the extracted hierarchy from the synthetic geometric shape data-set. ]{\includegraphics[width=0.9\textwidth]{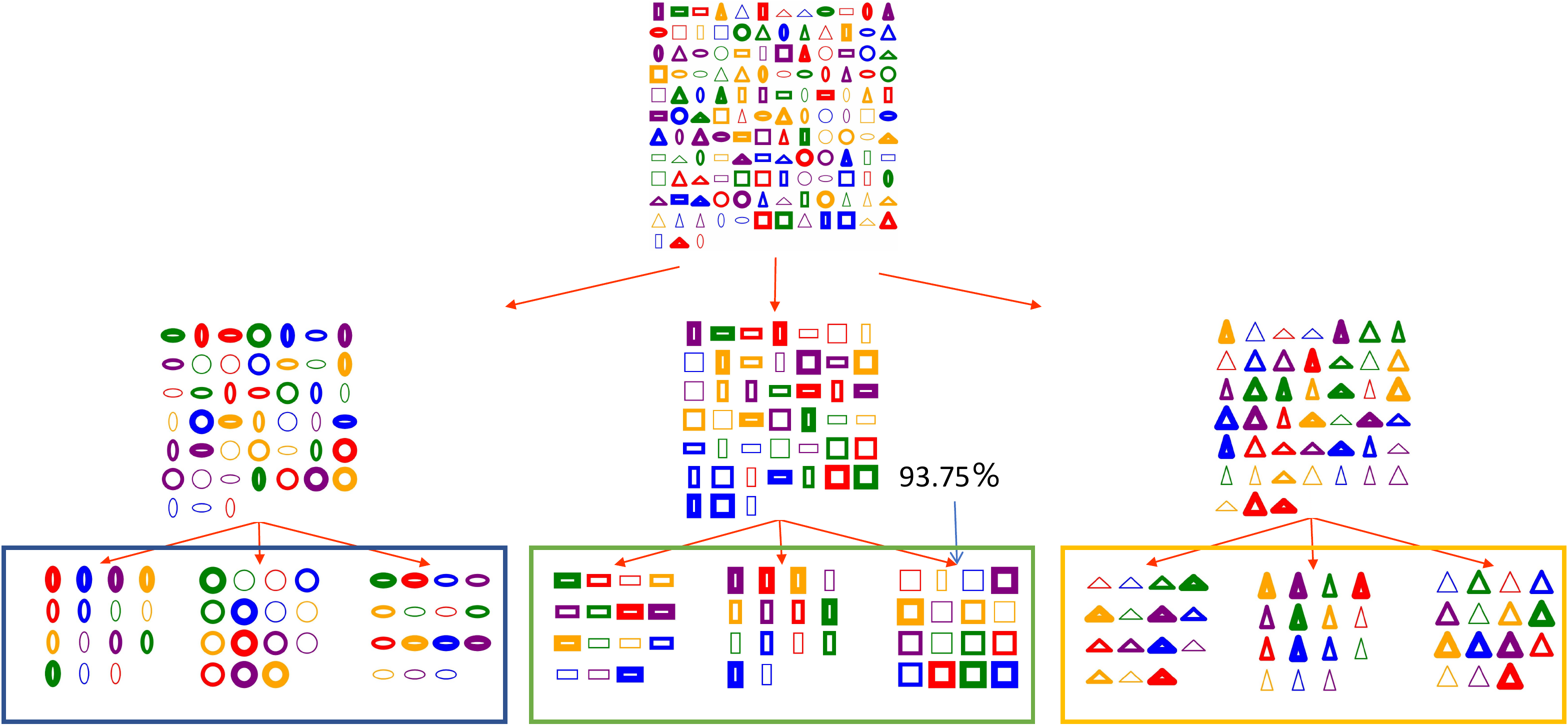}}\quad
		\subfloat[The last layer of the extracted hierarchy from the synthetic geometric shape data-set.  ]{\includegraphics[width=0.85\textwidth]{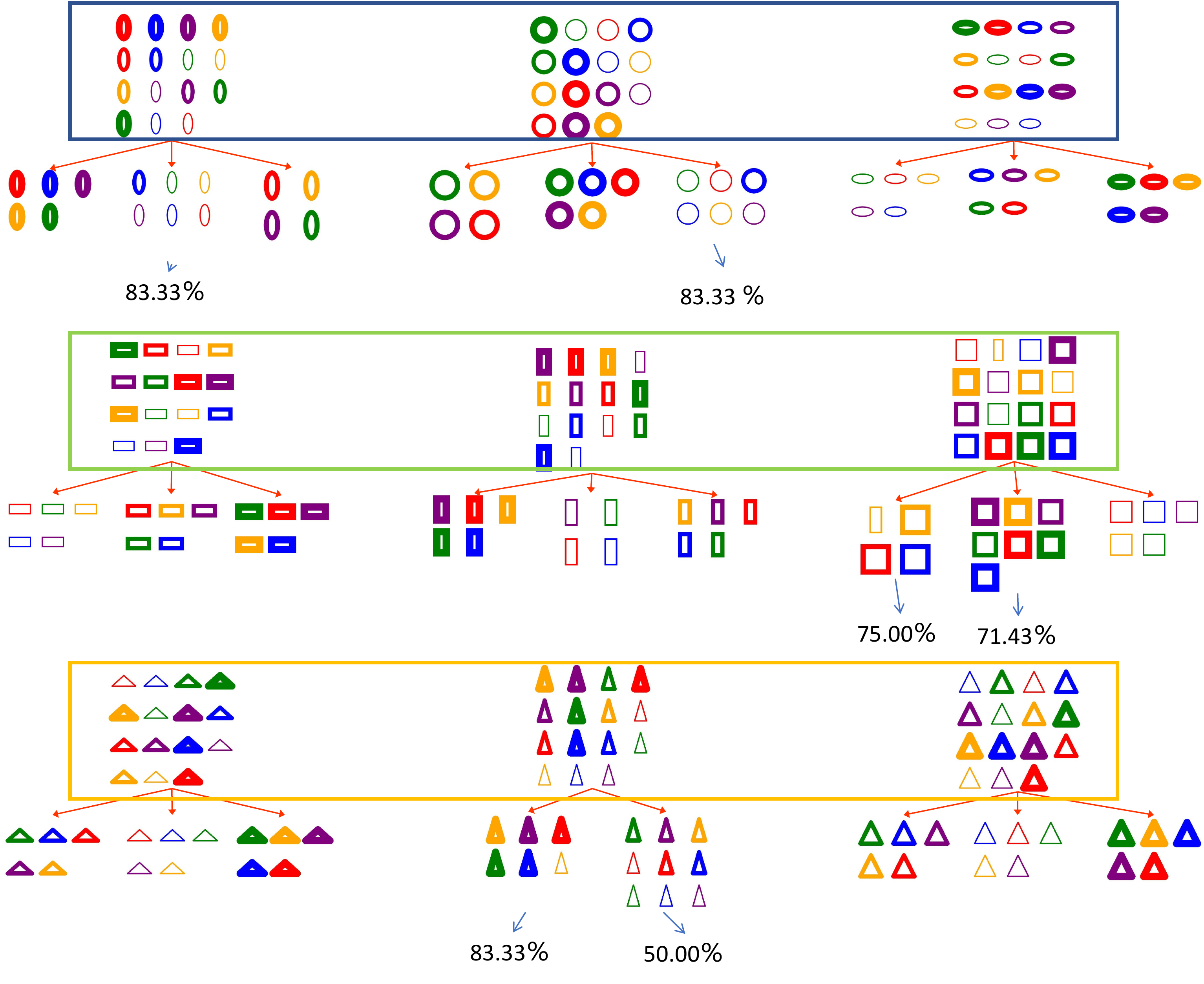}}\quad
		\caption{A four-layer hierarchy is extracted from the synthetic geometric shape data-set. The First 3 layers are shown in (a), and the last layer of the hierarchy is shown in (b) due to page size limitation. Clusters inside square-bound with the same color between (a) and (b) are the same. The first layer split into rectangles, circles, triangles three clusters, which is because the annotator thinks shape is the most discriminative property. In the second layer, shapes are again the discriminator; vertical stretched, horizontally stretched, and un-stretched images are cluster together. When the hierarchical tree goes down, we can notice that images begin to be organized based on different thicknesses, which are also shapes. Note only not perfectly pure nodes are marked with accuracy.}
		\label{simpleshape_H}
	\end{figure}

	The previous experiment has shown that our method is able to extract hierarchical structure by simulation. It is also crucial to test with real human responses . One challenge of real human involved experiments is how to evaluate our method since we don't know the latent hierarchical knowledge in advance. To deal with that, we test whether our method can uncover the well-studied shape bias in humans~\cite{landau1988importance} on a synthetic geometric data set. The data set contains triangles, circles, rectangles 3 different shapes. Each shape contains 3 deformations, 5 colors, and 3 thicknesses (See the first layer in Fig.~\ref{simpleshape_H}). Initially, we have the participant to answer 300 randomly chosen questions and train the model with a fixed margin of 0.2. During each following iteration, 300 actively sampled questions are answered. After 5 iterations, 1500 more triplets are collected to train the model with an adaptive margin.
	
	The extracted hierarchy are shown in Fig.~\ref{simpleshape_H}. Because most of the nodes have $100\%$ accuracy, we only mark the nodes which do not have perfect accuracies.  We also notice there are few images located in the wrong clusters. Remember, all the questions are answered by real humans, so mistaken answers are inevitable, and that makes the dendrogram not perfect.  The overall structure confirmed the human tend to believe shape is more important feature than other properties when they are asked to make choices.
	

	\subsection{Real Participant on Fashion-Mnist}
	Another real participant involved experiment is conducted on real-life image data set Fashion-Mnist~\cite{xiao2017fashion}. Accuracy can be calculated when the node of the hierarchy containing the original flat labels.  The ground truth latent perception is unknown before experiments and is discovered by our proposed method. We sample questions on a tiny Fashion-Mnist containing 900 images with 90 images for each class. During the first iteration, we randomly choose 600 questions, and train the model with a fixed margin of 0.4. In the following 2 iterations, we actively select 600 questions every iteration and train the model with an adaptive margin.
	
\begin{figure}[htbp] 
		
		\centering 
		\includegraphics[width=0.9\textwidth]{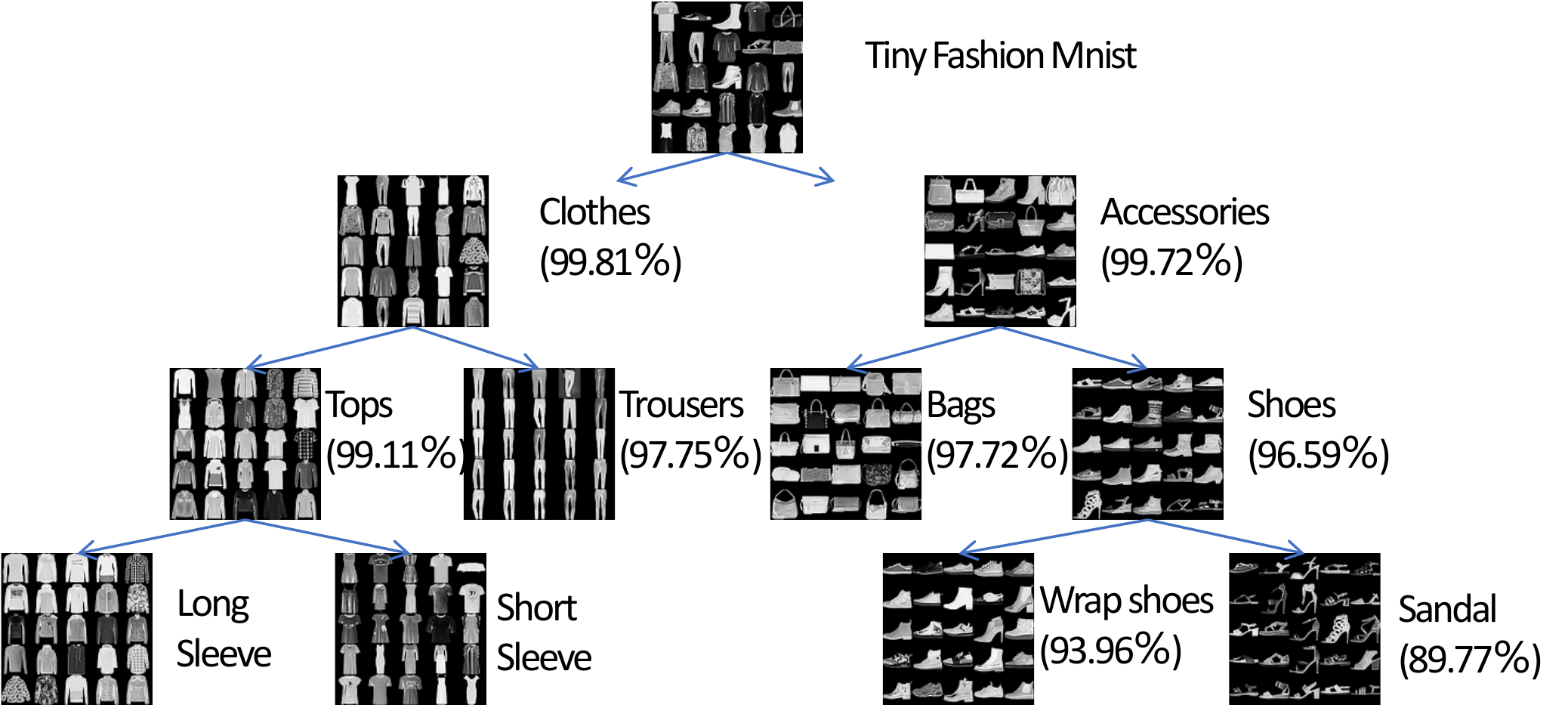} 
		\caption{Extracted hierarchy of tiny Fashion-Mnist.} 
		\label{FashionMnist_H}
	\end{figure}

	Fig.~\ref{FashionMnist_H} illustrates the extracted hierarchy. Note that Due to page size limitation, it is just a diagram, not the real hierarchical images. The first splits is based on cloth(left) and accessories(right) which further split down into tops (T-shirt, pullover, dress, coat, shirt)/trouser and  bags/shoes(sneaker, boot, sandal). Tops continue down into long sleeves/short sleeves, which are organized by the length of their sleeves regardless of the original cloth category. Further splits of shoes are wrap shoes(sneakers, boots) and sandals.  For the nodes composed of original Fashion-Mnist labels, we can calculate the accuracy of each node.
	
\begin{figure}[htbp] 
		\centering 
		\includegraphics[width=0.8\textwidth]{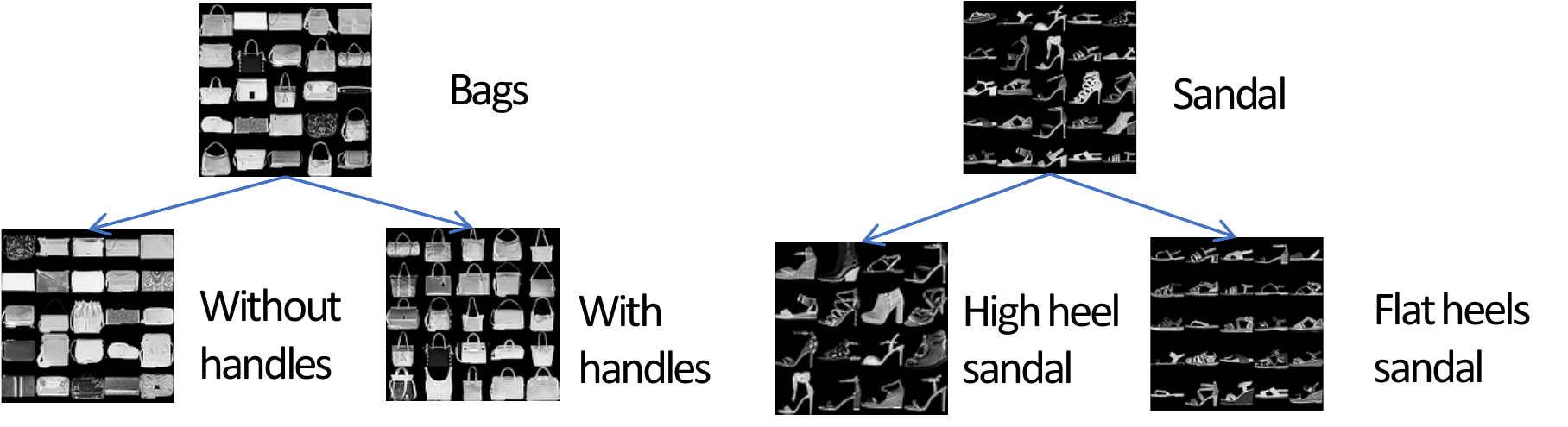} 
		\caption{Finer granularity of bags and sandals.} 
		\label{Finer}
	\end{figure}

	Further splits of sandals and bags can be seen in Fig.~\ref{Finer},  which presents us with a finer granularity than the original labels.

	\subsection{Comparison and Ablation Study}
\begin{figure}[htbp]
		\centering 
		
		\includegraphics[width=0.9\textwidth]{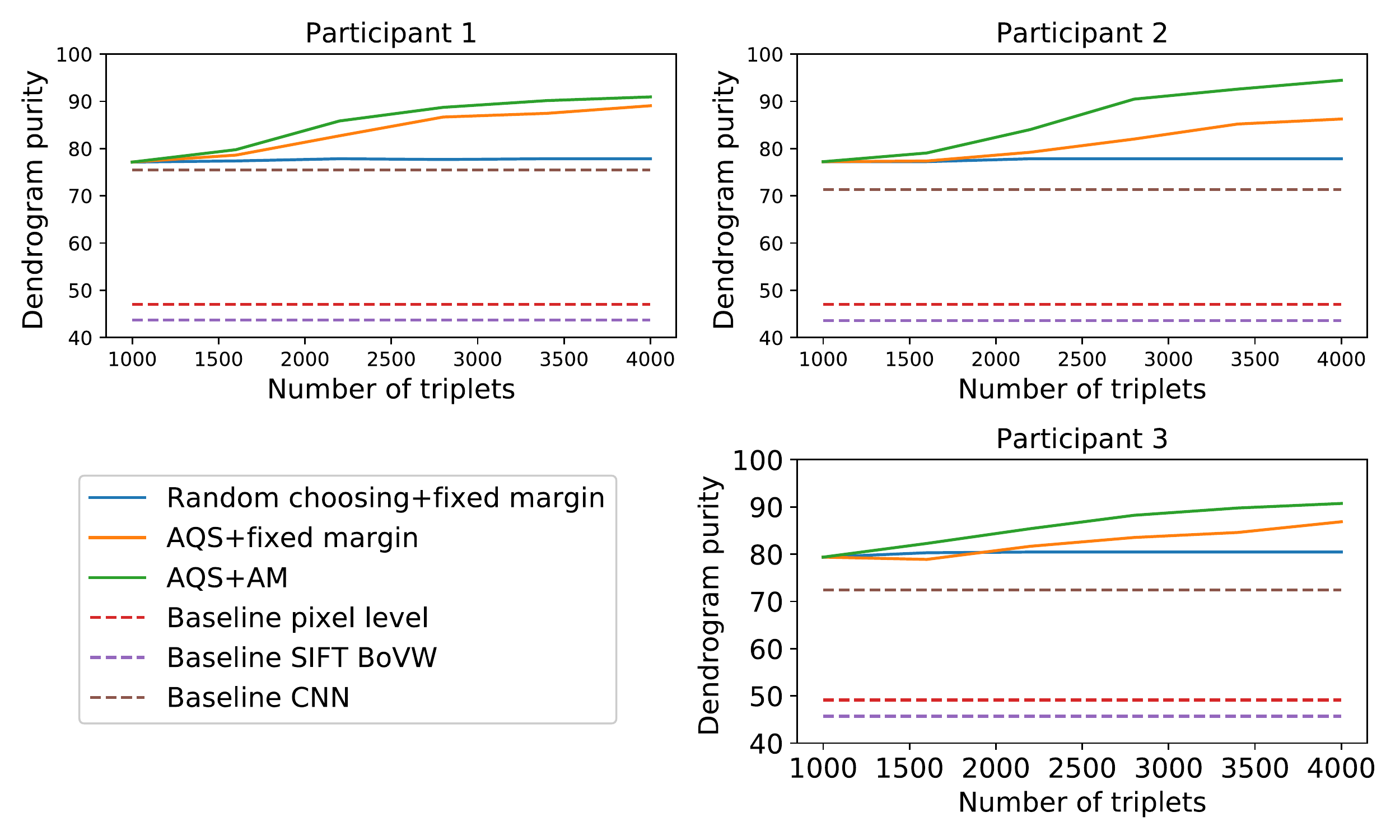} 
		\caption{Dendrogram purity training curves of different virtual responders. Our baselines are not trained with triplets,  so they keep flat. Even with minimum components, our proposed HKE has better performance than baselines. 
		} 
		\label{Alation_results}
	\end{figure}

	We compare our method with three baselines and perform an ablation study to evaluate our proposed components on tiny Cifar 10. After training with 4000 triplets, dendrogram purities are calculated in different experiment settings and reported in Table~\ref{Ablation_study}. Trained curves are shown in Fig.~\ref{Alation_results}.
	
	Three benchmarks are chosen for comparison. First,  pixel-level points are hierarchically clustered directly.   Then, SIFT features are extracted and represented as BoVW~\cite{yang2007evaluating} for hierarchical clustering. Besides, we train a CNN model with labels and hierarchically cluster on the last activation layer. For a fair comparison, we use the same CNN model as our HKE framework.  From Table~\ref{Ablation_study} and Fig.~\ref{Alation_results}, we can see the dendrogram purities of pixel-level and SIFT BoVW are lower than $50\%$. For baseline CNN, the dendrogram purity is higher than $70\%$.

	We also perform an ablation study to evaluate our proposed components. Experiment configuration are the same as in Section~\ref{sec_mutiple_cifar_10}.  In all the ablation experiments, dual-triplet is used to align with the 3AFC test. First, we sampling questions randomly and using a fixed margin of 0.4, the performance improves slightly in the first three iterations and keeps stuck even with more training data. Then, active question sampling(AQS) is applied to select questions with high-uncertainly and high utility, and performance gets a noticeable improvement. Finally, we integrate the adaptive margin(AM) and get the best performance.  
	
	When compare with HKE framework with baselines, we can notice that even with minimum components (Just 3AFC test, without AQS and AM), our proposed framework has better performance than all the baselines.
	

	\begin{table}[htbp]
		\newcommand{\tabincell}[2]{\begin{tabular}{@{}#1@{}}#2\end{tabular}}  
		\centering
		\caption{ Baselines and ablation study on Cifar 10 with 3 simulated participants}
		\begin{tabular}{|c|c|c|c|c|c|c|}  
			\hline
			participant &  \tabincell{c}{Base line\\ Pixel level} & \tabincell{c}{Base line\\SIFT BoVW}  & \tabincell{c}{Base line\\CNN} &\tabincell{c}{Random choosing\\+fixed margin}   & \tabincell{c}{AQS\\+fixed margin}  & \tabincell{c}{AQS\\ + AM}  \\
			\hline
			
			1 & 47.03 & 43.72& 75.5  &77.84 & 89.06 & \textcolor{red}{90.93}    \\ 
			2 & 46.99 & 43.57& 71.3  &77.84 & 86.25 & \textcolor{red}{94.44}    \\ 
			3 & 49.13 & 45.70& 72.4  &80.45 & 86.86 & \textcolor{red}{90.73}     \\ 
			
			\hline
		\end{tabular}
		\label{Ablation_study}
	\end{table}

	\section{Conclusion}
	
	Developing a machine learning formulation for a given task usually involves compromises that are related to how much and what kind of data is available, but also very much around the quantity and kind of expert knowledge that can be collected. The later tends to lead to oversimplifications, such as forcing a fixed set of labels and disregarding any more precise descriptions of the data. In this work, we present a method for knowledge elicitation from experts that can be applied to a broad set of machine learning formulations of problems. The results show that the proposed 3AFC testing with dual metric learning can extract latent hierarchical representations of data in a scalable way. This opens possibilities for applications in many domains where expert knowledge has a rich and complex underlying structure as in medical domain, biology as well as many industrial applications.

	%
	%
	\bibliographystyle{splncs04}
	\bibliography{ref}
	
\end{document}